\documentclass[11pt]{article}

\usepackage{fullpage}
\usepackage[ruled, linesnumbered, vlined, commentsnumbered]{algorithm2e}
\usepackage{graphicx,subfig} % figure related
\usepackage{amsfonts,amssymb,amsmath,amsthm,amsopn}	% math related
\usepackage{hhline,booktabs,colortbl,multirow,tabularx,threeparttable} % table related
\usepackage[listings,skins,breakable]{tcolorbox}

\usepackage{enumerate}
\usepackage{authblk}
\usepackage{footnote}
\usepackage{hyperref}
\usepackage{prettyref}
\usepackage{cite}
\usepackage{setspace}
\usepackage{color}
\usepackage{xcolor}  % Required for custom colors
\usepackage{scrpage2}
\usepackage{geometry}

\usepackage{tikz}
\usepackage{pgfplots}
\usetikzlibrary{positioning,shapes,shadows,arrows,calc}
\tikzstyle{component}=[rectangle, draw=black, rounded corners, fill=blue!40, drop shadow, text centered, anchor=north, text=white, minimum height=1cm]
\tikzstyle{arrow}=[->, thick]

\pgfplotsset{compat=1.12}
\usetikzlibrary{intersections}
\usetikzlibrary{pgfplots.statistics}
\usepgfplotslibrary{fillbetween}

\def\hlinew#1{%
  \noalign{\ifnum0=`}\fi\hrule \@height #1 \futurelet
   \reserved@a\@xhline}
\makeatother

\definecolor{myblue}{RGB}{34,31,217}
\definecolor{mycyan}{gray}{.7}
\definecolor{Gray}{gray}{0.9}

\newcommand{\pref}{\prettyref}

\newrefformat{fig}{Fig.~\ref{#1}}
\newrefformat{tab}{Table~\ref{#1}}
\newrefformat{sec}{Section~\ref{#1}}
\newrefformat{app}{Appendix~\ref{#1}}
\newrefformat{alg}{Algorithm~\ref{#1}}
\newrefformat{property}{Property~\ref{#1}}
\newrefformat{theorem}{Theorem~\ref{#1}}
\newrefformat{corollary}{Corollary~\ref{#1}}
\newrefformat{proposition}{Proposition~\ref{#1}}
\newrefformat{def}{Definition~\ref{#1}}
\newrefformat{eq}{equation~(\ref{#1})}

\title{\vspace{-1ex}\LARGE\textbf{Understanding the Automated Parameter Optimization on Transfer Learning for CPDP: An Empirical Study}\footnote{This manuscript is accepted for publication in ICSE 2020. The copyright of this paper has been permanently transferred to IEEE.}}

\author[1]{\normalsize Ke Li}
\author[2]{\normalsize Zilin Xiang}
\author[3]{\normalsize Tao Chen}
\author[1]{\normalsize Shuo Wang}
\author[4,5]{\normalsize Kay Chen Tan}
\affil[1]{\normalsize Department of Computer Science, University of Exeter, EX4 4QF, Exeter, UK}
\affil[2]{\normalsize College of Computer Science and Engineering, University of Electronic Science and Technology of China, 611731, Chengdu, China}
\affil[3]{\normalsize Department of Computer Science, Loughborough University, Loughborough, LE11 3TU, UK}
\affil[4]{\normalsize School of Computer Science, University of Birmingham, Birmingham, B15 2TT, UK}
\affil[5]{\normalsize Department of Computer Science, City University of Hong Kong, Tat Chee Avenue, Hong Kong}
\affil[$\ast$]{\normalsize Email: \texttt{k.li@exeter.ac.uk, zilin.xiang@gmail.com, t.t.chen@lboro.ac.uk}}
\date{}

\begin{document}
\maketitle

\vspace{-3ex}
{\normalsize\textbf{Abstract: }Data-driven defect prediction has become increasingly important in software engineering process. Since it is not uncommon that data from a software project is insufficient for training a reliable defect prediction model, transfer learning that borrows data/konwledge from other projects to facilitate the model building at the current project, namely cross-project defect prediction (CPDP), is naturally plausible. Most CPDP techniques involve two major steps, i.e., transfer learning and classification, each of which has at least one parameter to be tuned to achieve their optimal performance. This practice fits well with the purpose of automated parameter optimization. However, there is a lack of thorough understanding about what are the impacts of automated parameter optimization on various CPDP techniques. In this paper, we present the first empirical study that looks into such impacts on 62 CPDP techniques, 13 of which are chosen from the existing CPDP literature while the other 49 ones have not been explored before. We build defect prediction models over 20 real-world software projects that are of different scales and characteristics. Our findings demonstrate that: (1) Automated parameter optimization substantially improves the defect prediction performance of 77\% CPDP techniques with a manageable computational cost. Thus more efforts on this aspect are required in future CPDP studies. (2) Transfer learning is of ultimate importance in CPDP. Given a tight computational budget, it is more cost-effective to focus on optimizing the parameter configuration of transfer learning algorithms (3) The research on CPDP is far from mature where it is \lq not difficult\rq\ to find a better alternative by making a combination of existing transfer learning and classification techniques. This finding provides important insights about the future design of CPDP techniques.}
% in comparison with their non-preference-based baseline algorithms. In particular, the DM's preference information is elicited as a reference point, which represents her/his expectations on different objectives, in a priori manner. Experimental results demonstrate the usefulness of preference incorporation, especially for problems with many objectives. Moreover, we also find that the performance of different preference-based EMO algorithm can be largely influenced by their ways of how the preference information is used.}


\section{Introduction}
\label{sec:introdcution}

According to the latest \textit{Annual Software Fail Watch} report from Tricentis\footnote{https://www.tricentis.com/resources/software-fail-watch-5th-edition/}, software defects/failures affected over 3.7 billion people and caused \$1.7 trillion in lost revenue. In practice, stakeholders usually have limited software quality assurance resources to maintain a software project. Identifying high defect-prone software modules (e.g., files, classes or functions) by using advanced statistical and/or machine learning techniques, can be very helpful for software engineers and project managers to prioritize their actions in the software development life cycle and address those defects.

It is well known that a defect prediction model works well if it is trained with a sufficient amount of data~\cite{HallBBGC12}. However, it is controversial to obtain adequate data (or even having no data at all) in practice, especially when developing a brand new project or in a small company. By leveraging the prevalent transfer learning techniques~\cite{PanY10}, cross-project defect prediction (CPDP)~\cite{BriandMW02} has become increasingly popular as an effective way to deal with the shortage of training data~\cite{ZhouYLCLZQX18}. Generally speaking, its basic idea is to leverage the data from other projects (i.e., source domain projects) to build the model and apply it to predict the defects in the current one (i.e., target domain project).

Defect prediction models usually come with configurable and adaptable parameters (87\% prevalent classification techniques are with at least one parameter~\cite{Tantithamthavorn16,Tantithamthavorn19}), the settings of which largely influence the prediction accuracy when encountering unseen software projects~\cite{MendeK09,Mende10}. It is not difficult to envisage that the optimal settings for those parameters are problem dependent and are unknown beforehand. Without specific domain expertise, software engineers often train their defect prediction models with off-the-shelf parameters suggested in their original references. This practice may undermine the performance of defect prediction models~\cite{HallBBGC12} and be adverse to the research reproducibility of defect prediction studies~\cite{MittasA13,MenziesS12}. Recently, Tantithamthavorn \textit{et al.}~\cite{Tantithamthavorn16,Tantithamthavorn19} have empirically shown the effectiveness and importance of automated parameter optimization for improving the performance and stability of many prevalent classification techniques for defect prediction with manageable additional computational cost.

When considering CPDP, defect prediction become more complicated. To transfer knowledge from the source to the target domain, prevalent transfer learning techniques naturally bring additional configurable parameters. According to our preliminary literature study, 28 out of 33 most widely used CPDP techniques (85\%) require at least one parameter to setup in the transfer learning (or as known as domain adaptation) stage. This complication inevitably brings more challenges to the parameter optimization due to the further explosion of the parameter space, such as the extra difficulty of finding the optimal configuration and the increased computational cost for evaluating the model during optimization. Although hyper-parameter optimization (also known as automated machine learning) has been one of the hottest topics in the machine learning community~\cite{HKV2019}, to the best of our knowledge, little research have been conducted in the context of transfer learning.
% is an emerging topic in recent years, which automates the process of finding the optimal parameter settings with reasonable computation cost. to the best of our knowledge, little studies have been conducted %related to
% regarding whether the automated parameter optimization benefits the performance of transfer learning models, and none in the context of CPDP. 

Bearing these considerations in mind, in this paper, we seek to better understand how automated parameter optimization of transfer learning models can impact the performance in CPDP through a systematic and empirical study. In doing so, we aim to gain constructive insights based on which one can further advance this particular research area. To this end, the first research question (RQ) we wish to answer is:

%Further, more complex search space implies that more evaluations of the model become necessary during the optimization, which generates even larger computational cost. 
\vspace{0.5em}
\noindent
\framebox{\parbox{\dimexpr\linewidth-2\fboxsep-2\fboxrule}{
		\underline{\textbf{RQ1}}: \textbf{\textit{How much benefit of automated parameter optimization can bring to the performance of defect prediction models in the context of CPDP?}}
}}
\vspace{0.2em}

Answering \textbf{RQ1} is not straightforward, because transfer learning and classification are two intertwined parts in a CPDP model. Both of them have configurable parameters that can be used to adapt and control the characteristics of the CPDP model they produce. Therefore, the automated parameter optimization can be conducted by using three possible types of methods, all of which need to be studied for \textbf{RQ1}:
\begin{itemize}
    \item\underline{\texttt{Type-I}}: Naturally, it makes the most sense to optimize the parameters of both transfer learning and classification simultaneously. However, due to the large parameter space, it might be challenging to find the optimal configuration within a limited budget of computational cost.
    \item\underline{\texttt{Type-II}}: For the sake of taking the best use of computational cost, alternatively, parameters optimization may be conducted on one part of a CPDP model, i.e., either the transfer learning (denoted as \texttt{Type-II-1}) or the classification (denoted as \texttt{Type-II-2}), at a time; while the other part is trained by using the default parameters. In this way, the parameter space is reduced, and so does the necessary computational cost. However, this might not necessarily imply an undermined performance. For example, if transfer learning is the determinant part of a CPDP model in general, then optimizing it alone is expected to have at least the same level of result as optimizing both transfer learning and classification together (i.e., \texttt{Type I}) while causing much less computational cost.
    \item\underline{\texttt{Type-III}}: Finally, the automated parameter optimization can also be conducted in a sequential manner where the transfer learning part is optimized before the classification model\footnote{The parameters of a classification model is set as default values when optimizing the transfer learning part.}. In particular, each part is allocated half of the total budget of computational cost. In this method, although the total computational cost is exactly the same as that of \texttt{Type I}, the parameter space is further constrained, which enables more focused search behaviors.
\end{itemize}

The above also motivates our second RQ, in which we ask:

\vspace{0.5em}
\noindent
\framebox{\parbox{\dimexpr\linewidth-2\fboxsep-2\fboxrule}{
		\underline{\textbf{RQ2}}: \textbf{\textit{What is the most cost effective type of automated parameter optimization given a limited amount of computational cost?}}
}}
\vspace{0.2em}

Investigating \textbf{RQ1} and \textbf{RQ2} would inevitably require us to go through a plethora of transfer learning and classification techniques proposed in the machine learning literature~\cite{PanY10}. During the process, we found that the transfer learning and classification techniques in existing CPDP models are either selected in a problem-specific or ad-hoc manner. Little is known about the versatility of their combinations across various CPDP tasks with different characteristics. Because of such, our final RQ seeks to understand:

\vspace{0.5em}
\noindent
\framebox{\parbox{\dimexpr\linewidth-2\fboxsep-2\fboxrule}{
		\underline{\textbf{RQ3}}: \textbf{\textit{Whether the state-of-the-art combinations of transfer learning and classification techniques can indeed represent the generally optimal settings?}}
}}
\vspace{0.2em}

To address the above RQs, we apply \texttt{Hyperopt}~\cite{BergstraYC13}, an off-the-shelf hyper-parameter optimization toolkit\footnote{http://hyperopt.github.io/hyperopt/}, as the fundamental optimizer on the CPDP models considered in our empirical study. Comprehensive and empirical study is conducted on 62 combinations of the transfer learning algorithms and classifiers, leading to a total of 37 different parameters to be tuned, and using 20 datasets from real-world open source software projects. In a nutshell, our findings answer the RQs as below:

\begin{itemize}
    \item[--]\textbf{To RQ1:} Automated parameter optimization can substantially improve the CPDP techniques considered in our empirical study. In particular, 77\% of the improvements have been classified as \textit{huge} according to the Cohen's $d$ effect size.
    \item[--]\textbf{To RQ2:} Transfer learning is the most determinant part in CPDP while optimizing its parameters alone can achieve better CPDP performance than the other types of automated parameter optimization.
    \item[--]\textbf{To RQ3:} No. Some \lq newly\rq\ developed CPDP techniques, with under-explored combinations of transfer learning and classification techniques, can achieve better (or at least similar) performance than those state-of-the-art CPDP techniques.
\end{itemize}

Drawing on those answers, our empirical study, for the first time, provides new insights that help to further advance this field of research\footnote{To enable a reproducible research, all the experimental data and source code of our empirical study can be found at \url{https://github.com/COLA-Laboratory/icse2020/}.}:
\begin{itemize}
    \item Automated parameter optimization can indeed provide significant improvement to the CPDP model, within which optimizing the parameters of transfer learning is the most determinant part. This observation suggests that future research on the optimizer can primarily focus on this aspect in the design and hence prevent wasting efforts on other methods that provide no better performance but generating extra computational cost only.
    \item The state-of-the-art combinations of transfer learning and classifier are far from being optimal, implying that the selection of combination is at least as important as the parameter tuning. As a result, future work should target a whole portfolio of optimization, tuning not only the parameters, but also the algorithmic components, i.e., the selection of appropriate transfer learning and classifier pair, of a CPDP model.
\end{itemize}

The rest of this paper is organized as follows. \pref{sec:setup} provides the methodology used to conduct our empirical study. \pref{sec:results} present and analyze the experimental results. Thereafter, the results and threats to validity are discussed in~\pref{sec:discussion} along with a pragmatic literature review in~\pref{sec:related work}. At the end, \pref{sec:conclusions} concludes the findings in this paper and provides some potential future directions.

\section{The Empirical Study Methodology}
\label{sec:setup}

This section elaborates the methodology and experimental setup of our empirical study, including the dataset selection, the working principle of \texttt{Hyperopt}, the system architecture of automated parameter optimization for CPDP model building and the performance metric used to evaluate the performance of a CPDP model.

\subsection{Dataset Selection}
\label{sec:data}

In our empirical study, we use the following six criteria to select the datasets for CPDP model building.
\begin{enumerate}
    \item To promote the research reproducibility of our experiments, we choose datasets hosted in public repositories.
    \item To mitigate potential conclusion bias, the datasets are chosen from various corpora and domains. More importantly, the shortlisted datasets in our empirical study have been widely used and tested in the CPDP literature.
    \item If the dataset has more than one version, only the latest version is used to train a CPDP model. This is because different versions of the same project share a lot of similarities which simplify transfer learning tasks.
    \item To avoid overfiting a CPDP model, the datasets should have enough instances for model training.
    \item To promote the robustness of our experiments, it is better to inspect the datasets to rule out data that are either repeated or having missing values.
    \item To resemble real-world scenarios, it is better to choose datasets from open source projects provided by iconic companies.
\end{enumerate}

According to the first two criteria and some recent survey papers on the CPDP research~\cite{Herbold17,HerboldTG18,ZhouYLCLZQX18,HosseiniTG19}, we shortlist five publicly available datasets (\textit{i.e.,} \texttt{JURECZKO}, \texttt{NASA}, \texttt{SOFTLAB}, \texttt{AEEEM}, \texttt{ReLink}). Note that these datasets are from different domains and have been frequently used in the literature. To meet the fourth criterion, we further rule out \texttt{SOFTLAB} while \texttt{NASA} is also not considered in our experiments due to its relatively poor data quality~\cite{ShepperdSSM13} according to the fifth criterion. In summary, the datasets used in our experiments consist of 20 open source projects with 10,952 instances. The characteristics of each dataset are summarized as follows:

\begin{itemize}
    \item[--] \texttt{AEEEM}~\cite{DAmbrosLR10}: This dataset contains 5 open source projects with 5,371 instances. In particular, each instance has 61 metrics with two different types, including static and process metrics like the entropy of code changes and source code chorn.
    \item[--] \texttt{ReLink}~\cite{WuZKC11}: This dataset consists of 3 open source projects with 649 instances. In particular, each instance is with 26 static metrics. Note that the defect labels are further manually verified after being generated from source code management system commit comments.
    \item[--] \texttt{JURECZKO}~\cite{JureczkoM10}: This dataset originally consists of 92 released products collected from open source, proprietary and academic projects. To meet the first criterion, those proprietary projects are ruled out from our consideration. To meet the last criterion, the projects merely for academic use are excluded from \texttt{JURECZKO}. Moreover, since the projects in \texttt{JURECZKO} have been updated more than one time, according to the third criterion, only the latest version is considered in our experiments. At the end, we choose 12 open source projects with 4,932 instances in our empirical study.
\end{itemize}

\begin{figure*}[t!]
    \centering
	\includegraphics[width=.7\linewidth]{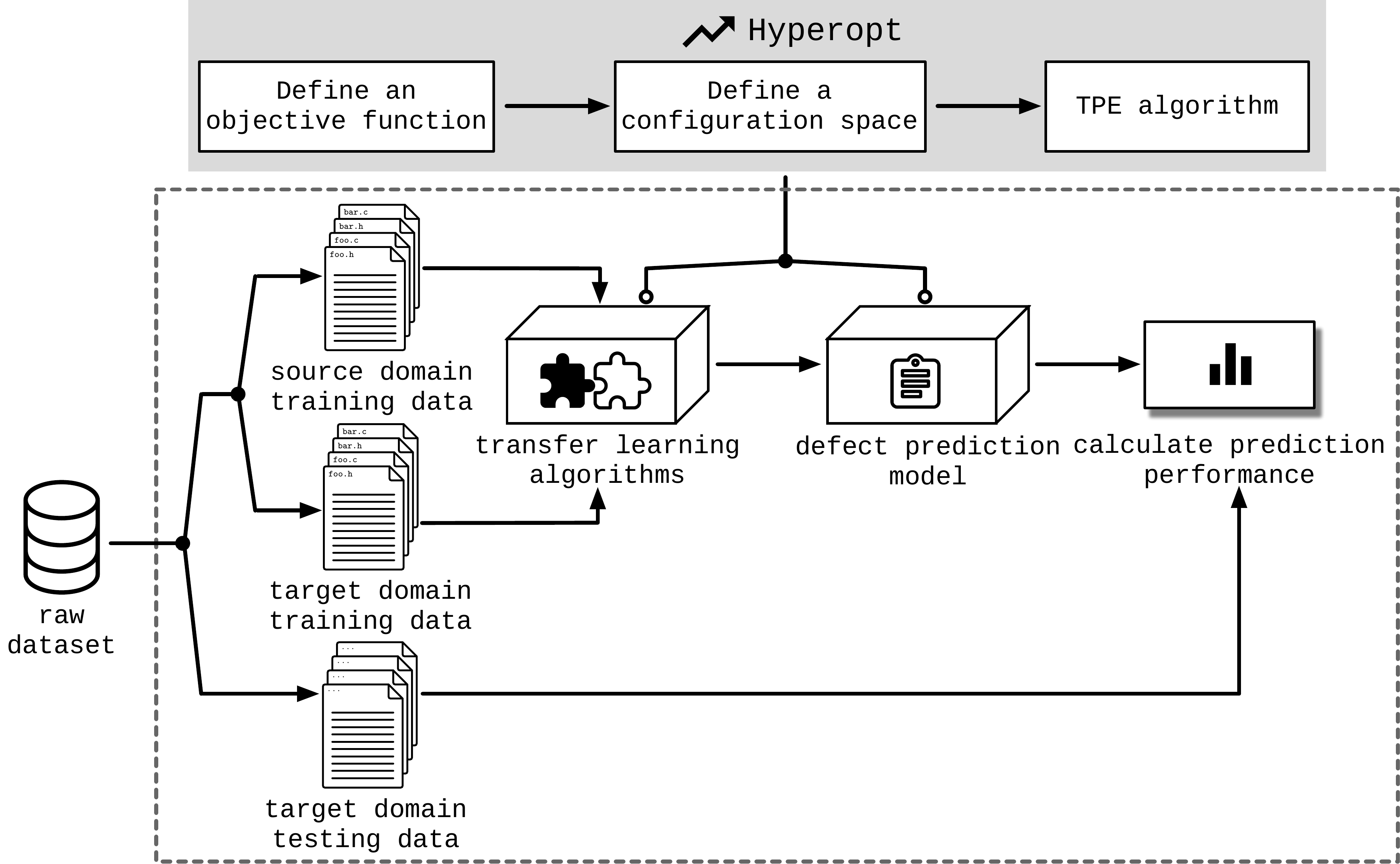}
	\caption{The architecture of automated parameter optimization on CPDP model by using Hyperopt.}
	\label{fig:system_CPDP}
\end{figure*}

\subsection{\texttt{Hyperopt} for Automated Parameter Optimization}
\label{sec:hyperopt}

\texttt{Hyperopt}\footnote{http://hyperopt.github.io/hyperopt/} is a Python library that provides algorithms and the software infrastructure to optimize hyperparameters of machine learning algorithms. \texttt{Hyperopt} uses its basic optimization driver  \texttt{hyperopt.fmin} to optimize the parameter configurations of the CPDP techniques considered in our empirical study. Using \texttt{Hyperopt} requires three steps. %The way to use \texttt{Hyperopt} is only made up of three steps.
\begin{itemize}
    \item[--] \textbf{\underline{Define an objective function}:} As suggested in~\cite{BergstraYC13a}, \texttt{Hyperopt} provides a flexible level of complexity when specifying an objective function that takes inputs and returns a loss users want to minimize. In our experiments, the inputs are parameters associated with the transfer learning and classification techniques as shown in the third column of ~\pref{tab:transfer_learning} and~\pref{tab:classification} respectively.
    \item[--] \textbf{\underline{Define a configuration space}:} The configuration space in the context of \texttt{Hyperopt} describes the domain over which users want to search. The last column of~\pref{tab:transfer_learning} and~\pref{tab:classification} list the configuration space of the corresponding parameter.
    \item[--] \textbf{\underline{Choose an optimization algorithm}:} \texttt{Hyperopt} provides two alternatives, \textit{i.e.,} random search~\cite{BergstraB12} and Tree-of-Parzen-Estimators (TPE) algorithm~\cite{BergstraBBK11}, to carry out the parameter optimization. In our experiments, we choose to use the TPE algorithm because the sequential model-based optimization has been recognized as one of the most effective hyperparameter optimization methods in the auto-machine learning literature~\cite{FeurerH19}. In practice, \texttt{Hyperopt} provides a simple interface to deploy the TPE algorithm where users only need to pass \texttt{algo=hyperopt.tpe.suggest} as a keyword argument to \texttt{hyperopt.fmin}.
\end{itemize}

\subsection{Architecture for Optimizing CPDP Model}
\label{sec:architecture}

\pref{fig:system_CPDP} shows the architecture of using \texttt{Hyperopt} to optimize the performance of CPDP models. As shown in~\pref{fig:system_CPDP}, the implementation steps of our empirical study are given below.
\begin{enumerate}
  %part of the target domain data is used to train the CPDP model while the remaining ones are used for the testing purpose.
    \item Given a raw dataset with $N$ projects, $N-1$ of which are used as the source domain data while the other project is used as the target domain data. In particular, all source domain data is used for training whilst the data from the target domain is used for testing. To mitigate a potentially biased conclusion on the CPDP ability, all 20 projects will be used as target domain data in turn during our empirical study.
    \item The CPDP model building process consists of two intertwined parts, \textit{i.e.,} transfer learning and defect prediction model building. 
    \begin{itemize}
        \item Transfer learning aims to augment data from different domains by selecting relevant instances or assigning appropriate weights to different instances, \textit{etc}. \pref{tab:transfer_learning} outlines %some important information 
        the parameters of the transfer learning techniques considered in our empirical study.
        \item Based on the adapted data, many off-the-shelf classification techniques can be applied to build the defect prediction model. \pref{tab:classification} outlines 
        %some important information
        the parameters of the classification techniques considered in this paper.
    \end{itemize}
    \item The performance of the defect prediction ability of the CPDP model is at the end evaluated upon the hold-out set from the target domain data.
\end{enumerate}

\begin{table*}[htbp]
	\scriptsize
	\centering
	\caption{Parameters of the transfer learning techniques considered in our experiments}
	\begin{threeparttable}
	\scalebox{1}{
	\begin{tabular}{l|lp{23.5em}l}
		\hline
		\multicolumn{1}{c|}{\multirow{1}[4]{*}{\textbf{Transfer learning}}} & \multicolumn{3}{c}{\textbf{Parameters}} \\
		\cline{2-4}    \multicolumn{1}{c|}{} & \textbf{Name} & \textbf{Description} & \textbf{Range} \\
		\hline
		Bruakfilter & k     & The number of neighbors to each point (default=10) \tnote{[N]}& [1, 100] \\
		\hline
		\multirow{2}[2]{*}{DS} & topN  &  The number of closest training sets (default=5) \tnote{[N]}& [1, 15] \\
		& FSS   & The ratio of unstable features filtered out (default=0.2) \tnote{[R]}& [0.1, 0.5] \\
		\hline
		\multirow{2}[2]{*}{DSBF} & Toprank & The number assigned to 1 when performing feature reduction (default=1) \tnote{[N]}& [1, 10] \\
		& k     & The number of neighbors to each point (default=25) \tnote{[N]}& [1, 100] \\
		\hline
		\multirow{4}[2]{*}{TCA} & kernel & The type of kernel (default=\lq linear\rq) \tnote{[C]}& \{\lq primal\rq, \lq linear\rq, \lq rbf\rq, \lq sam\rq\} \\
		& dimension & The dimension after tranforing (default=5) \tnote{[N]}& [5, max(N\_source, N\_target)] \\
		& lamb  & Lambda value in equation (default=1) \tnote{[R]}& [0.000001, 100] \\
		& gamma & Kernel bandwidth for \lq rbf\rq\ kernel (default=1) \tnote{[R]}& [0.00001, 100] \\
		\hline
		\multirow{2}[2]{*}{DBSCANfilter} & eps   & The maximum distance between two samples for one to be considered as in the neighborhood of the other (default=1.0) \tnote{[R]}& [0.1, 100] \\
		& min\_samples & The number of samples (or total weight) in a neighborhood for a point to be considered as a core point (default=10) \tnote{[N]}& [1, 100] \\
		\hline
		\multirow{2}[2]{*}{Universal} & p-value &  The associated p-value for statistical test (default=0.05) \tnote{[R]}& [0.01, 0.1] \\
		& quantifyType &The type of quantifying the difference between distributions (default=\lq cliff\rq) \tnote{[C]}& \{\lq cliff\rq, \lq cohen\rq\} \\
		\hline
		\multirow{2}[2]{*}{DTB} & k     &  The number of neighbors to each point (default=10) \tnote{[N]}& [1, 50] \\
		& T     &  The maximum number of iterations (default=20) \tnote{[N]}& [5, 30] \\
		\hline
		Peterfilter & r     & The number of points in each cluster (default=10) \tnote{[N]}& [1, 100] \\
		\hline
	\end{tabular}}	
	\begin{tablenotes}
		\item[{[N]}] An integer value from range
		\item[{[R]}] A real value from range
		\item[{[C]}] A choice from categories
	\end{tablenotes}
	\end{threeparttable}
	\label{tab:transfer_learning}%
\end{table*}%

\begin{table*}[htbp]
	\scriptsize
	\centering
	\caption{Parameters of the classification techniques considered in our experiments}
	\begin{threeparttable}
	\scalebox{0.8}{
	\begin{tabular}{p{18.11em}|p{9.055em}p{22.445em}p{15.29em}}
		\hline
		\multirow{1}[4]{*}{\centering \textbf{Classification techniques}} & \multicolumn{3}{c}{\textbf{Parameters}} \\
		\cline{2-4}    \multicolumn{1}{c|}{} & \textbf{Name} & \textbf{Description} & \textbf{Range} \\
		\hline
		K-Nearest Neighbor (KNN) & n\_neighbors &  The number of neighbors to each point (default=1) \tnote{[N]} & [1, 50] \\
		\hline
		\multirow{2}[2]{*}{Boost} & n\_estimators & The maximum number of estimators (default=50) \tnote{[N]} & [10, 1000] \\
		\multicolumn{1}{l|}{} & learning rate &  A factor that shrinks the contribution of each classifier (default=1) \tnote{[R]} & [0.01, 10] \\
		\hline
		\multirow{4}[10]{*}{Classification and Regression Tree (CART)} & criterion & The maximum number of estimators (default=10) \tnote{[N]} & [10, 100] \\
		\multicolumn{1}{l|}{} & max\_features & The function to measure the quality of a split (default=\lq gini\rq) \tnote{[C]} & \{\lq gini\rq, \lq entropy\rq\} \\
		\multicolumn{1}{l|}{} & splitter & The number of features to consider when looking for the best split (default=\lq auto\rq) \tnote{[C]} & \{\lq auto\rq, \lq sqrt\rq, \lq log2\rq\} \\
		\multicolumn{1}{l|}{} & min\_samples\_split & The minimum number of samples required to split an internal node (default=2) \tnote{[N]}& [2, N\_source/10] \\
		\hline
		\multirow{4}[10]{*}{Random Forest (RF)} & n\_estimators & The maximum number of estimators (default=10) \tnote{[N]} & [10, 100] \\
		\multicolumn{1}{l|}{} & criterion & The function to measure the quality of a split (default=\lq gini\rq) \tnote{[C]}& \{\lq gini\rq, \lq entropy\rq\} \\
		\multicolumn{1}{l|}{} & max\_features & The number of features to consider when looking for the best split (default=\lq auto\rq) \tnote{[C]}& \{\lq auto\rq, \lq sqrt\rq, \lq log2\rq\} \\
		\multicolumn{1}{l|}{} & min\_samples\_split & The minimum number of samples required to split an internal node (default=2) \tnote{[N]} & [2, N\_source/10] \\
		\hline
		\multirow{4}[8]{*}{Support Vector Machine (SVM)} & kernel &  The type of kernel (default=\lq poly\rq) \tnote{[C]}& \{\lq rbf\rq, \lq linear\rq, \lq poly\rq, \lq sigmoid\rq\} \\
		\multicolumn{1}{l|}{} & degree &  Degree of the polynomial kernel function (default=3) \tnote{[N]}& [1, 5] \\
		\multicolumn{1}{l|}{} & coef0 & Independent term in kernel function. It is only significant in \lq poly\rq\ and \lq sigmoid\rq\ (default=0.0) \tnote{[R]}& [0, 10] \\
		\multicolumn{1}{l|}{} & gamma & Kernel coefficient for \lq rbf\rq, \lq poly\rq\ and \lq sigmoid\rq\ (default=1) \tnote{[R]}& [0.01, 100] \\
		\hline
		\multirow{3}[2]{*}{Multi-layer Perceptron (MLP)} & active & Activation function for the hidden layer (default=\lq relu\rq) \tnote{[C]} & \{\lq identity\rq, \lq logistic\rq, \lq tanh\rq, \lq relu\rq\} \\
		\multicolumn{1}{l|}{} & alpha & L2 penalty (regularization term) parameter (default=0.0001) \tnote{[R]}& [0.000001, 1] \\
		\multicolumn{1}{l|}{} & iter  & Maximum number of iterations (default=200) \tnote{[N]}& [10, 500] \\
		\hline
		\multirow{2}[2]{*}{Ridge} & alpha &Regularization strength (default=1) \tnote{[R]}& [0.0001, 1000] \\
		\multicolumn{1}{l|}{} & normalize & Whether to standardize (default=\lq False\rq) \tnote{[C]}& \{\lq True\rq, \lq False\rq\} \\
		\hline
		Naive Bayes (NB) & NBType & The type of prior distribution (default=\lq Gaussian\rq) \tnote{[C]}& \{\lq gaussian\rq, \lq multinomial\rq, \lq bernoulli\rq \} \\
		\hline
	\end{tabular}}
	\begin{tablenotes}
		\item[{[N]}] An integer value from range
		\item[{[R]}] A real value from range
		\item[{[C]}] A choice from categories
	\end{tablenotes}
	\end{threeparttable}
	\label{tab:classification}
\end{table*}%

\begin{table}[htbp]
\small
\centering
\caption{Overview of existing CPDP techniques considered in our empirical study.}
%\resizebox{\columnwidth}{!}{
\begin{threeparttable}

\begin{tabular}{c|c||c|c}
\hline
\textbf{CPDP Techniques} & \textbf{Reference} & \textbf{CPDP Techniques} & \textbf{Reference} \\ \hline
Bruakfilter (NB)                  &   \cite{TurhanMBS09}                 & DS+BF (RF)                        & \multirow{2}{*}{\cite{AmasakiKY15}}  \\ \cline{1-3}
Petersfilter (RF)                 & \multirow{3}{*}{\cite{PetersMM13}}  & DS+BF (NB)                        &                    \\ \cline{1-1} \cline{3-4} 
Petersfilter (NB)                 &                    & DTB                               & \cite{ChenFST15}                   \\ \cline{1-1} \cline{3-4} 
Petersfilter (KNN)                &                    & DBSCANfilter (RF)                 & \multirow{3}{*}{\cite{7336025}}  \\ \cline{1-3}
FSS+Bagging (RF)                  & \multirow{2}{*}{\cite{HePMY13}}  & DBSCANfilter (NB)                 &                    \\ \cline{1-1} \cline{3-3}
FSS+Bagging (NB)                  &                    & DBSCANfilter (KNN)                &                    \\ \hline
UM                                & \cite{ZhangMKZ16}                   &                                   &                    \\ \hline
\end{tabular}

\begin{tablenotes}
\item[] \footnotesize The classifier is shown in the brackets while outside part is the transfer learning technique. UM uses Universal to carry out transfer learning and Naive Bayes as a classifier. DTB uses DTB to carry out transfer learning part and Naive Bayes to conduct classification.
\end{tablenotes}
\end{threeparttable}
\label{tab:existing_CPDP}
\end{table}

Note that there are 13 CPDP techniques considered in our empirical study. All of them are either recognized as the state-of-the-art in the CPDP community or used as the baseline for many other follow-up CPDP techniques. \pref{tab:existing_CPDP} lists the combination of transfer learning and classifier used in each CPDP technique. To carry out the automated parameter optimization for a CPDP technique, \texttt{Hyperopt} is allocated 1,000 function evaluations. In our context, one function evaluation represents the complete model training process of a CPDP technique with a trial parameter setup, which can be computationally expensive. To carry out statistical analysis over our experiments, the optimization over each CPDP technique is repeated 10 times.

\subsection{Performance Metric}
\label{sec:metric}

To evaluate the performance of different CPDP methods for identifying defect-prone modules, we choose the area under the receiver operator characteristic curve (AUC) in our empirical study. This is because AUC is the most widely used performance metric in the defect prediction literature. In addition, there are two distinctive characteristics of AUC: 1) different from other prevalent metrics like precision and recall, the calculation of AUC does not depend on any threshold~\cite{ZhouYLCLZQX18} which is difficult to tweak in order to carry out an unbiased assessment; and 2) it is not sensitive to imbalanced data which is not unusual in defect prediction~\cite{LiJZ18}. Note that the larger the AUC values, the better prediction accuracy of the underlying classification technique is. In particular, the AUC value ranges from 0 to 1 where 0 indicates the worst performance, 0.5 corresponds a random guessing performance and 1 represents the best performance.

\section{Results and Analysis}
\label{sec:results}

In this section, we will present the experimental results of our empirical study and address the research questions posed in~\pref{sec:introdcution}.

\subsection{On the Impacts of Automated Parameter Optimization Over CPDP Techniques}
\label{sec:RQ1}

\subsubsection{Research Method}

To address \textbf{RQ1}, we investigate the magnitude of AUC performance improvement achieved by the CPDP model optimized by \texttt{Hyperopt} versus the one trained by its default parameter setting. Under a target domain (project), instead of comparing the difference of vanilla AUC values\footnote{To make our results self-contained, the vanilla AUC values are given in the supplementary document of this paper and can be found in \url{https://github.com/COLA-Laboratory/icse2020}} for all repeated runs, we use Cohen's $d$ effect size~\cite{Cohen88} to quantify such magnitude. This is because it is simple to calculate and has been predominately used as the metric for automated parameter optimization of defect prediction model~\cite{Tantithamthavorn16}. Given two sets of samples say $S_1$ and $S_2$, Cohen's $d$ effect size aims to provide a statistical measure of the standardized mean difference between them:
\begin{equation}
    d=\frac{\mu_1-\mu_2}{s},
\end{equation}
where $\mu_1$ and $\mu_2$ is the mean of $S_1$ and $S_2$ respectively; $s$ is as defined as the pooled standard deviation:
\begin{equation}
    s=\sqrt{\frac{(n_1-1)s_1^2+(n_2-1)s_2^2}{n_1+n_2-2}}
\end{equation}
where $n_1$ and $n_2$ is the number of samples in $S_1$ and $S_2$ respectively; while $s_1$ and $s_2$ are the corresponding standard deviations of the two sample sets. To have a conceptual assessment of the magnitude, according to the suggestions in~\cite{Sawilowsky09}, $d<0.2$ is defined as \textit{negligible}, $0.2<d\leq0.5$ is treated as \textit{small}, $0.5<d\leq0.8$ is regarded as \textit{medium}, $0.8<d\leq1.0$ is large while it is \textit{huge} if $d$ goes beyond 1.0.

As introduced in~\pref{sec:introdcution}, we run four different optimization types (as presented in~\pref{sec:introdcution}) in parallel upon each baseline CPDP technique. To investigate whether \texttt{Hyperopt} can improve the performance of a CPDP technique, we only present the results from the best optimization type to make our conclusion sharper. For each CPDP technique, we use a violin plot to show the distributions of its median values of Cohen's $d$ effect size obtained by optimizing the parameters of this CPDP techniques on 20 projects. To have a better understanding of the effect of automated parameter optimization upon different CPDP techniques, the violin plots are sorted, from left to right, by the median values of Cohen's $d$ effect size in a descending order.

\subsubsection{Results}

\begin{figure}[t!]
    \centering
	\includegraphics[width=.5\linewidth]{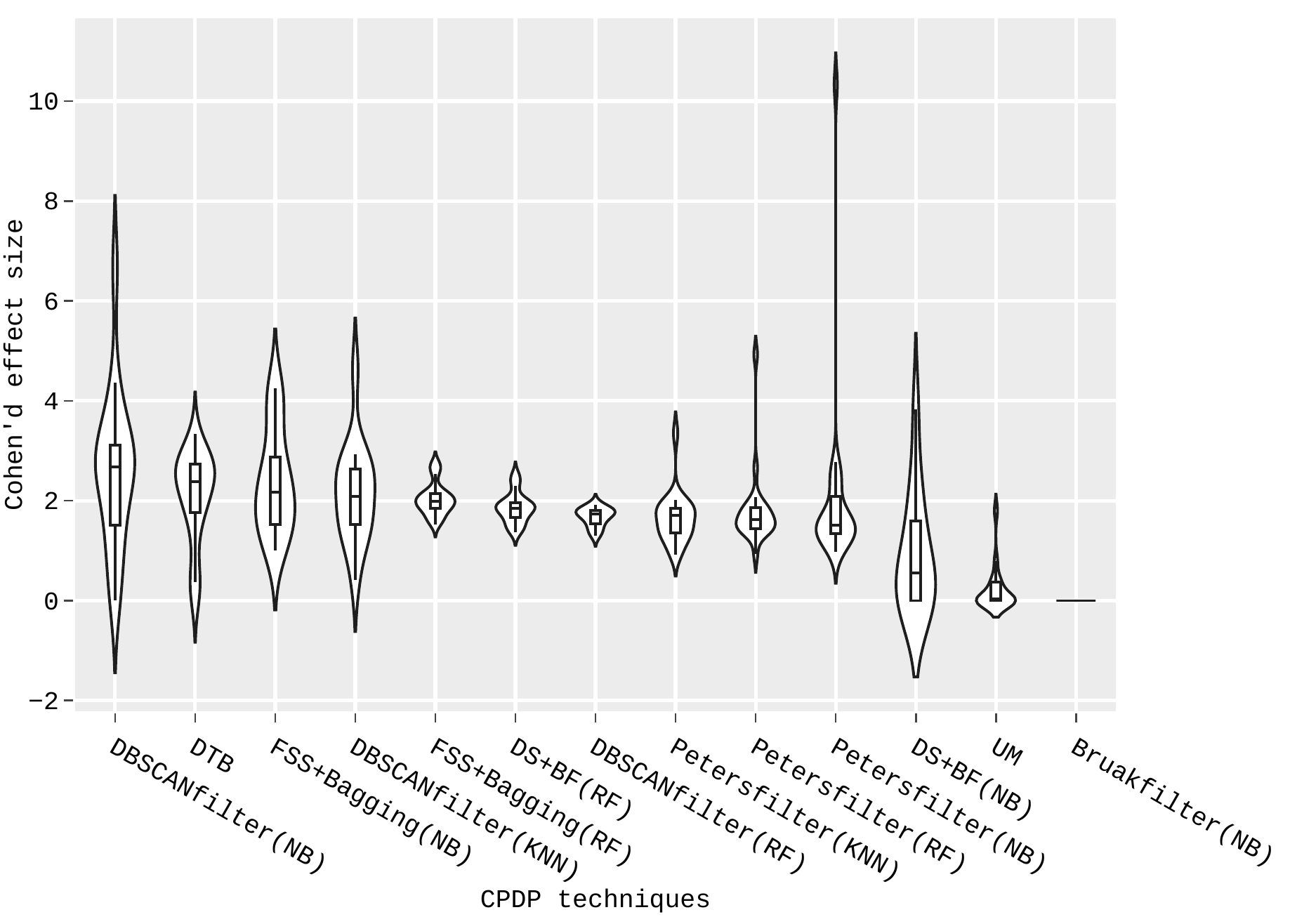}
	\caption{The AUC performance improvement in terms of Cohen's $d$ effect size for each studied CPDP technique.}
	\label{fig:RQ1}
\end{figure}

From the comparison results shown in~\pref{fig:RQ1}, we clearly see that the performance of 12 out of 13 (around 92\%) existing CPDP techniques have been improved after automated parameter optimization. In addition, according to the categorization in~\cite{Sawilowsky09}, the magnitudes of most AUC performance improvements are substantial and important. In particular, ten of them (around 77\%) are classified as \textit{huge}; while the performance improvements achieved by optimizing the parameters of DS+BF (NB) belong to the \textit{medium} scale. In particular, \texttt{Hyperopt} leads to the most significant performance improvement on DBSCANfilter (NB) and DTB. On the other hand, the magnitudes of AUC performance improvement achieved by optimizing the parameters of UM and Bruakfilter (NB) are \textit{negligible} (i.e., Cohen's $d<0.2$). It is worth mentioning that \texttt{Hyperopt} cannot improve the performance of Bruakfilter (NB) any further at all considered projects. This might suggest that the original parameter configuration of Bruakfilter (NB) is already optimal. Overall, we obtain the following findings:

% \vspace{-1em}
\noindent
\framebox{\parbox{\dimexpr\linewidth-2\fboxsep-2\fboxrule}{
		\underline{\textbf{Answer to RQ1}}: \textbf{\textit{Automated parameter optimization can improve the performance of defect prediction models in the context of CPDP. In particular, the performance of 10 out of 13 (around 77\%) studied CPDP techniques have been improved substantially (i.e., huge in terms of Cohen's $d$ effect size value).}}
}}

\subsection{Comparing Different Types of Parameter Optimization}
\label{sec:RQ2}

\subsubsection{Research Method}

To answer \textbf{RQ2}, we investigate the performance of four different types of parameter optimization, as introduced in~\pref{sec:introdcution}. To have an overview of the result, for each CPDP technique, we record the best parameter optimization type. In addition, for each optimization type, we also record the number of times that its AUC value is significantly better than the other peers over all CPDP technique and project pairs. In addition, to have a better intuition on the effect of different types of parameter optimization over each CPDP technique, we use violin plots to show the distributions of the median values of Cohen's $d$ effect size obtained over 20 projects\footnote{Again, to make our results self-contained, the vanilla AUC values are given in the supplementary document of this paper and can be found in \url{https://github.com/COLA-Laboratory/icse2020}}.

\subsubsection{Results}

\begin{figure}[t!]
    \centering
	\includegraphics[width=.5\linewidth]{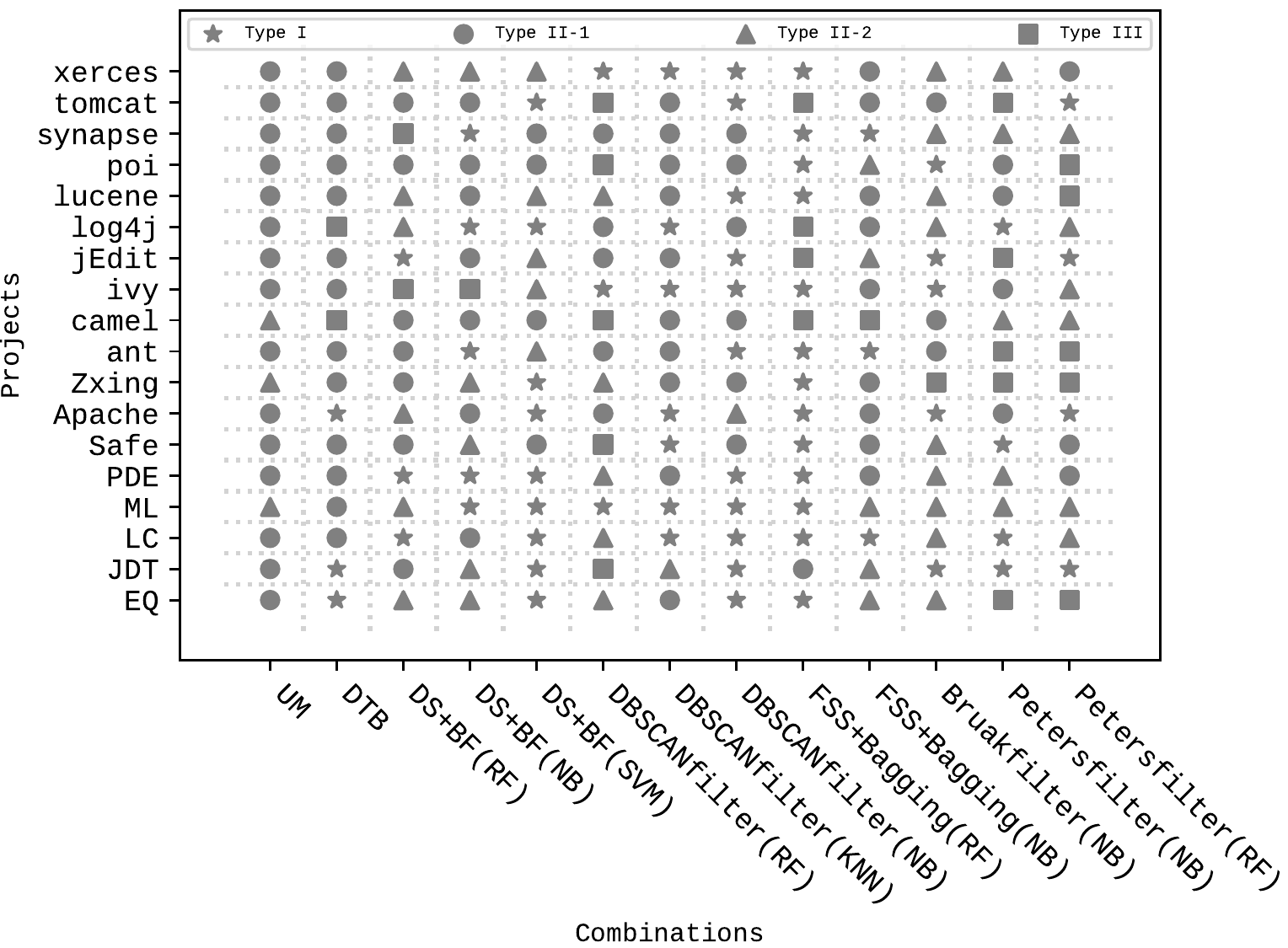}
	\caption{Distribution of the best parameter optimization type for each CPDP technique and project pair.}
	\label{fig:RQ2-distribution}
\end{figure}

\begin{figure}[t!]
    \centering
	\includegraphics[width=.5\linewidth]{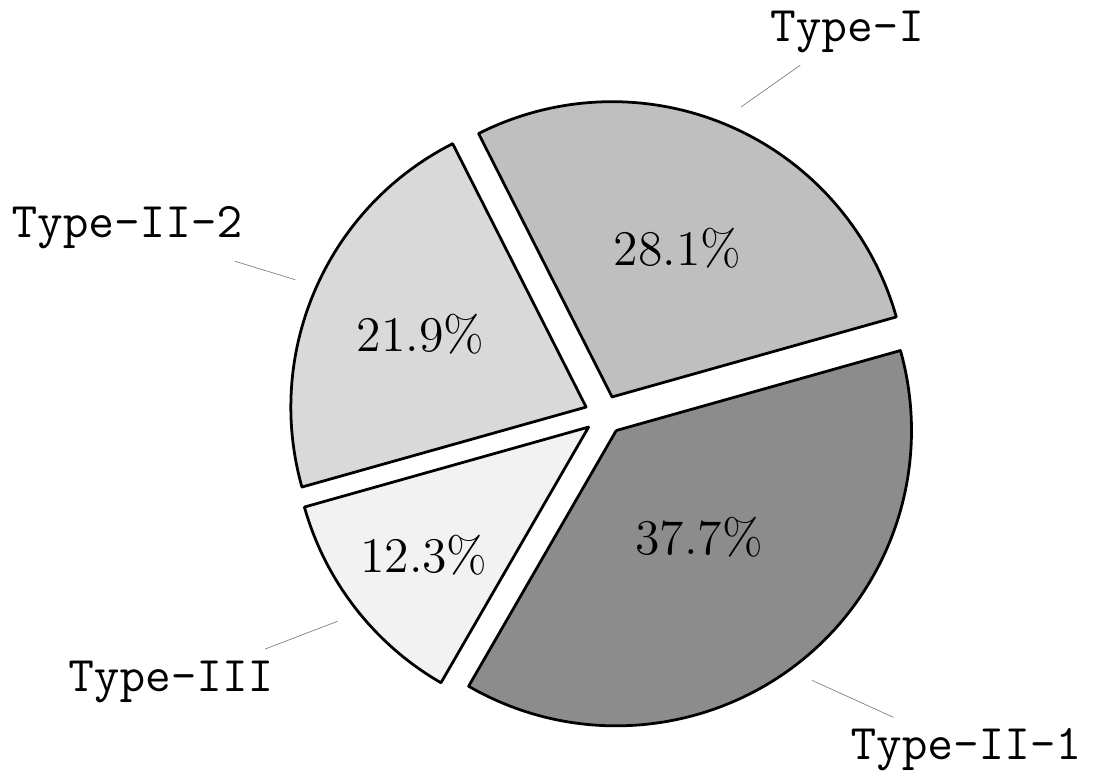}
	\caption{The percentage of significantly better performance achieved by different optimization types.}
	\label{fig:RQ2-pie}
\end{figure}

\begin{figure*}[htbp]
    \centering
	\includegraphics[width=\linewidth]{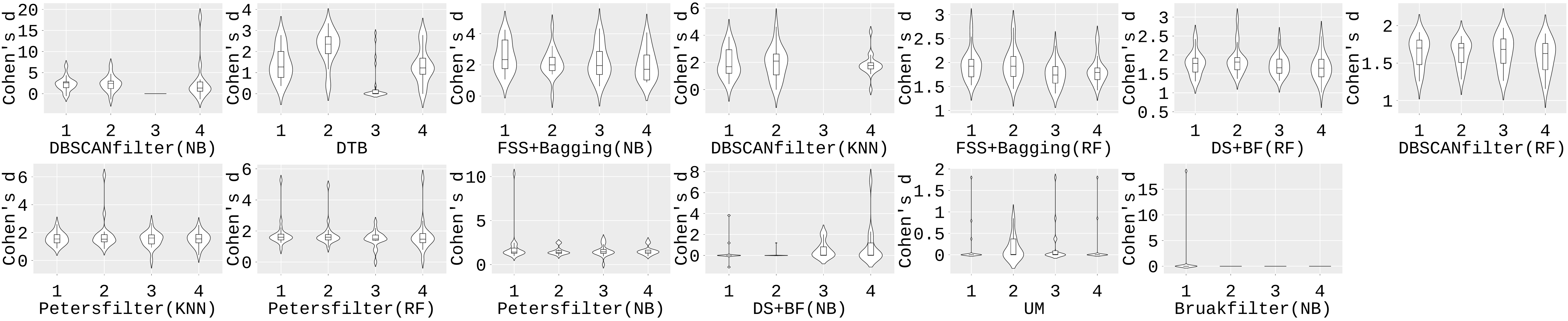}
	\caption{Violin plots of AUC performance improvement in terms of Cohen's $d$ effect size for each studied CPDP technique. 1: \texttt{Type-I}, 2: \texttt{Type-II-1}, 3: \texttt{Type-II-2}, 4: \texttt{Type-III}}
	\label{fig:RQ2-d}
\end{figure*}

From the results shown in~\pref{fig:RQ2-distribution}, we have a general impression that \texttt{Type-II-1} plays as the best parameter optimization type in most cases. In particular, for UM and DTB, \texttt{Type-II-1} almost dominates the list. The second best parameter optimization type is \texttt{Type-I} whilst the worst one is \texttt{Type-III} which is rarely ranked as the best optimization type in most cases.

The pie chart shown in~\pref{fig:RQ2-pie} is a more integrated way to summerize the results collected from~\pref{fig:RQ2-distribution}. From this figure, we can see that the type of only optimizing the parameters associated with the transfer learning part in CPDP is indeed more likely to produce the best performance. In particular, 37.7\% of the best AUC performance is achieved by \texttt{Type-II-1}. It is even better than simultaneously optimizing the parameters of both transfer learning and classification parts, i.e., \texttt{Type-I}, which wins on 28.1\% comparisons. This might be because given the same amount of computational budget, simultaneously optimizing the parameters of both transfer learning and classification parts is very challenging due to the large search space. As a result, \texttt{Hyperopt} might run out of function evaluations before approaching the optimum on neither part. On the other hand, if \texttt{Hyperopt} only focuses on optimizing the parameters of the transfer learning part, the search space is significantly reduced. Therefore, although only part of the parameters is considered, it is more likely to find the optimal parameter configuration of the transfer learning part within the limited number of function evaluations. The same applied to \texttt{Type-II-2}, which only focus on optimizing the parameters of the classification techniques. However, as shown in~\pref{fig:RQ2-pie}, the performance of \texttt{Type-II-2} is not as competitive as \texttt{Type-I} and \texttt{Type-II-1}, implying that the classification part is less important than the transfer learning part in CPDP, which eventually obscures the benefit brought by the reduced search space. Finally, we see that sequentially optimizing the transfer learning and classification parts with equal budget of computation is the worst optimization type (\texttt{Type-III}). This is because it does not only fail to fully optimize both parts before exhausting the function evaluations, but also ignore the tentative coupling relationship between the parameters associated with both the transfer learning and classification.

From the results shown in~\pref{fig:RQ2-d}, we find that the performance difference between different types of parameter optimization is not very significant in most CPDP techniques. Nonetheless, we can still observe the superiority of \texttt{Type-I} and \texttt{Type-II-1} over the other two optimization types in most performance comparisons. In particular, for DBSCANfilter (NB), DBSCANfilter (KNN) and DTB, only optimizing the parameters of the classification part does not make any contribution to the performance improvement on the corresponding CPDP techniques. This observation is also aligned with our previous conclusion that the transfer learning part is more determinant than the classification part in CPDP. In summary, we find that:

\vspace{0.5em}
\noindent
\framebox{\parbox{\dimexpr\linewidth-2\fboxsep-2\fboxrule}{
		\underline{\textbf{Answer to RQ2}}: \textbf{\textit{Given a limited amount of computational budget, it is more plausible to focus on the parameter optimization of the transfer learning part in CPDP than the other types, including optimizing the configurations of both transfer learning and classification simultaneously. This observation also demonstrates that the transfer learning part is more determinant in CPDP.}}
}}

\subsection{Comparing Different Combinations of Transfer Learning and Classification Techniques for CPDP}

\subsubsection{Research Method}

To address \textbf{RQ3}, we build and compare 62 different CPDP models by combining those transfer learning and classification techniques listed in~\pref{tab:transfer_learning} and~\pref{tab:classification} respectively. 13 out of these 62 combinations exist in the literature. The remaining 49 combinations can be regarded as \lq new\rq\ CPDP techniques. Because DTB requires to update the weights of its training instances during the training process, it cannot work with KNN or MLP which do not support online training data adjustments. In other words, the combinations DTB-KNN and DTB-MLP are ruled out from our empirical study. For a better visualization, instead of presenting the performance of all 62 combinations together, we only show the 10 best CPDP techniques. We use violin plots to show the distributions of their AUC values.

In addition, for each project, we compare the performance of the best CPDP technique from the existing literature and those \lq newly\rq\ developed in this paper. To have a statistically sound conclusion, we apply the widely used Wilcoxon signed-rank sum test with a 0.05 significance level to validate the statistical significance of those comparisons~\cite{LiZZL09,LiZLZL09,LiFK11,LiKWCR12,CaoKWL12,LiKCLZS12,LiKWTM13,LiWKC13,LiK14,CaoKWL14,LiZKLW14,LiFKZ14,CaoKWLLK15,WuKZLWL15,LiDZK15,LiKZD15,LiKD15,LiDZ15,LiDZZ17,WuKJLZ17,WuLKZZ17,LiDAY17,LiDY18,ChenLY18,WuLKZ20,ChenLBY18,KumarBCLB18,LiCFY19,WuLKZZ19,LiCSY19,Li19,BillingsleyLMMG19,ZouJYZZL19,LiuLC19,LiZ19,LiXT19,GaoNL19}. In particular, if the best candidate from the \lq newly\rq\ developed CPDP techniques is significantly better than the best one from the current literature, it is denoted as \textit{win}; if their difference is not statistically significant, it is denoted as \textit{tie}; otherwise, it is denoted as \textit{loss}. We keep a record of the number of times of these three scenarios.

\begin{figure*}[htbp]
    \centering
    \includegraphics[width=\linewidth]{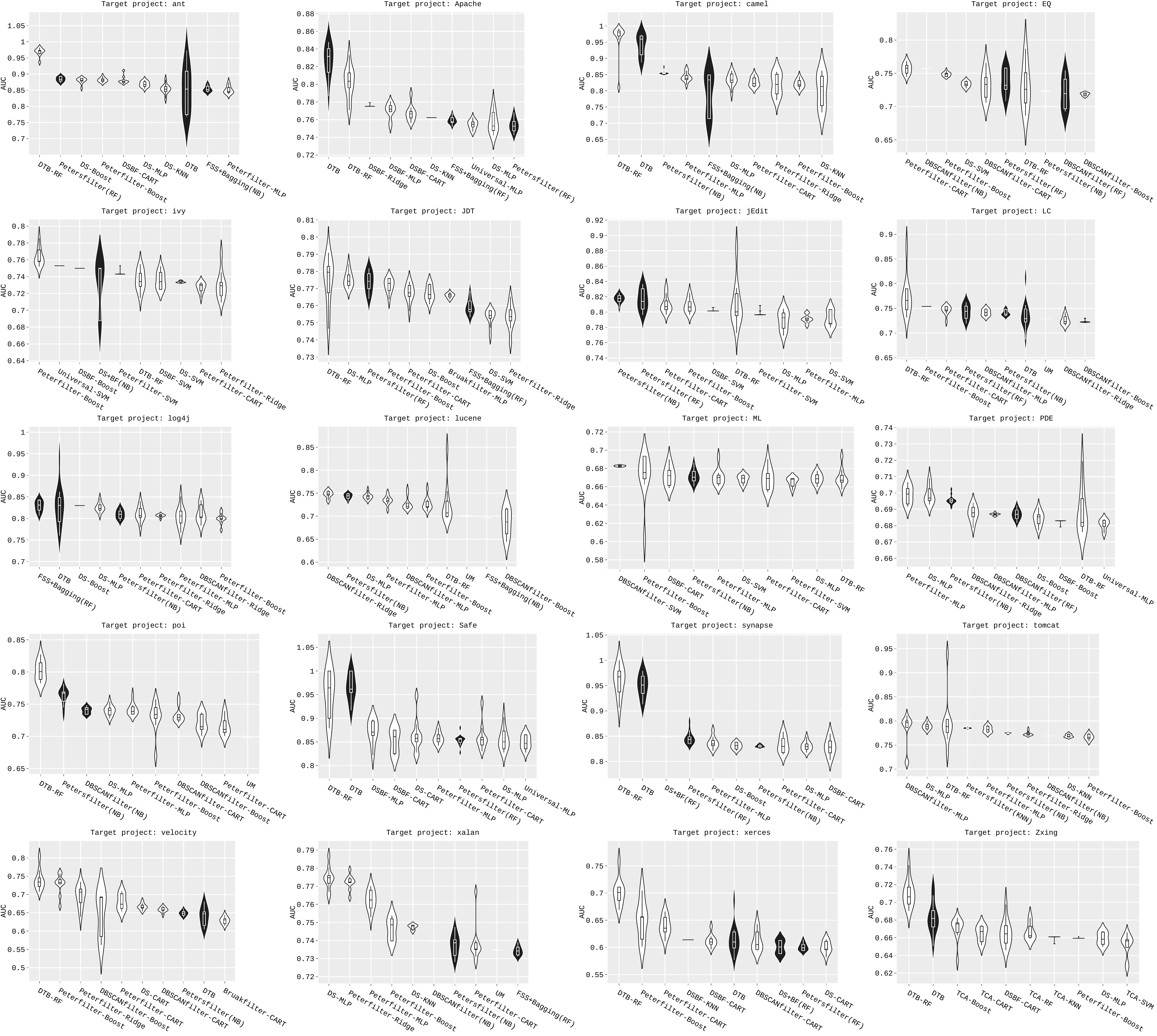}
    \caption{Violin plots of AUC values obtained by top 10 CPDP techniques for different projects. In particular, existing CPDP techniques are with black charts whilst \lq newly\rq\ developed ones are with white .}
    \label{fig:RQ3-d}
\end{figure*}

\subsubsection{Results}

From the violin plots shown in~\pref{fig:RQ3-d}, we find that the list of top 10 CPDP techniques varies from different projects. In particular, DTB-RF is the best CPDP technique as it appears in all top 10 lists and is ranked as the first place in 9 out of 20 projects. Furthermore, we notice that DTB, Peterfilter and DBSCANfilter are the best transfer learning techniques for CPDP because they were used as the transfer learning part in the CPDP techniques of all top 10 lists. From these observations, we conclude that CPDP techniques also follow the \textit{no-free-lunch theorem}~\cite{DolpertM97}. In other words, there is no universal CPDP technique capable of handling all CPDP tasks of the data have different characteristics.

\begin{figure}[t!]
	\centering
	\includegraphics[width=.5\linewidth]{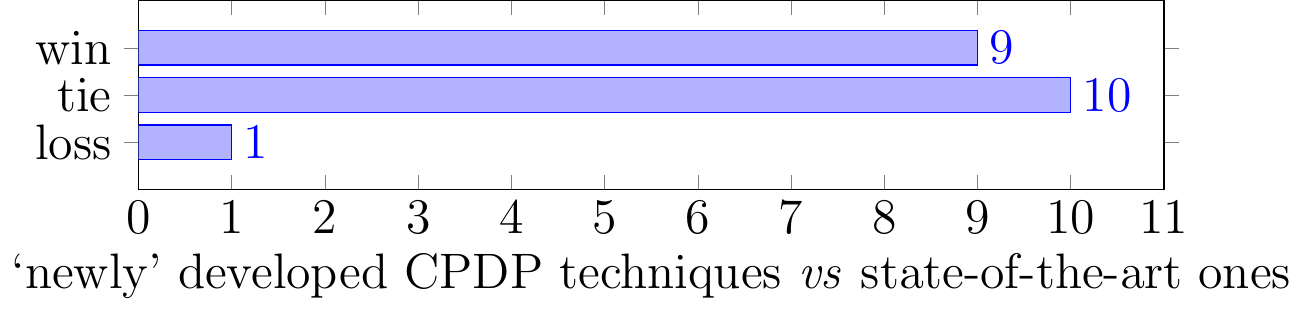}
	\caption{Comparisons of state-of-the-art CPDP techniques and those \lq newly\rq\ developed in this paper.}
	\label{fig:RQ3_bar}
\end{figure}

\pref{fig:RQ3_bar} gives the statistics of the comparison between the best CPDP technique from the existing literature and the one from our \lq newly\rq\ developed portfolio. From the statistics, we can see that the CPDP techniques newly developed in this paper, by making a novel combination of transfer learning and classification techniques, are better than those existing ones in the literature. Remarkably, they are only outperformed by the existing CPDP techniques under one occasion. From this observation, we envisage that the current research on CPDP is far from mature. There is no rule-of-thumb a practitioner can follow to design a CPDP technique for the black-box dataset at hand. For \textbf{RQ3}, we have the following findings:

\vspace{0.5em}
\noindent
\framebox{\parbox{\dimexpr\linewidth-2\fboxsep-2\fboxrule}{
		\underline{\textbf{Answer to RQ3}}: \textbf{\textit{The current research on CPDP techniques is far from mature. Given a CPDP task, there is no rule-of-thumb available for a practitioner to follow in order to 1) design an effective technique that carries out an appropriate transfer learning and classification; and 2) find out the optimal configurations of their associated parameters.}}
}}

\section{Discussions}
\label{sec:discussion}

\subsection{Insights Learned from Our Empirical Study}

Our empirical study, for the first time, reveals some important and interesting facts that provide new insights to further advance the research on CPDP. 

Through our comprehensive experiments, we have shown that automated parameter optimization can significantly improve the defect prediction performance of various CPDP techniques with a \textit{huge} effect size in general. In particular, it is surprising but also exciting to realize that optimizing the parameters of transfer learning part only is generally the most cost effective way of tuning the defect prediction performance of CPDP techniques. Such a finding offers important insight and guidance for future research: given a limited amount of computational budget, designing sophisticated optimizer for the parameters of CPDP techniques can start off by solely considering the transfer learning part, without compromising the performance.

Our other insightful finding is to reveal the fact that the current research on CPDP is far from mature. In particular, many state-of-the-art combinations of transfer learning and classification techniques are flawed, and that the best combination can be case dependent. This suggests that automatically finding the optimal combination of CPDP techniques for a case is a vital research direction, and more importantly, the combination should be tuned with respect to the optimization of parameters. Such an observation can derive a brand new direction of research, namely the portfolio optimization of transfer learning and classifier that lead to an automated design of CPDP technique.

\subsection{Threats to Validity}

Similar to many empirical studies in software engineering, our work is subject to threats to validity. Specifically, internal threats can be related to the number of function evaluations used in the optimization. Indeed, a larger amount of function evaluations may lead to better results in some cases. However, the function evaluation in our empirical study is expensive and time consuming, as every single one needs to go through the full machine learning training process, validation and testing. As a result, a proper choice should be a good trade-off between the performance and time. To mitigate this threat, we have run numbers of options in a trial-and-error manner. We then concluded that 1,000 function evaluations is deemed as a balanced choice without compromising the validity of our conclusions. Furthermore, to mitigate bias, we repeated 10 times for each CPDP technique on a project, which is acceptable considering the cost of function evaluation.

Construct threats can be raised from the selected quality indicator. In this work, AUC has been chosen as the key performance indicator in our empirical comparisons. The selection is mainly driven by its simplicity (no extra parameter is required) and its robustness (insensitive to imbalanced data). In addition, AUC has been widely recognised as one of the most reliable performance indicator in the machine learning community~\cite{LingHZ03}. The significance of differences have also been assessed in terms of the effect size using Cohen's $d$.

Finally, external threats are concerned with the dataset and CPDP techniques studied. To mitigate such, we have studied 62 CPDP techniques, including 13 combinations from existing work on CPDP and 49 other combinations that are new to the CPDP community but widely applied in classic machine learning research. Further, as discussed in~\pref{sec:data}, our studied dataset covers a wide spectrum of the real-world defected projects with diverse nature, each of which was selected based on six systematic criteria. Such a tailored setting, although not exhaustive, is not uncommon in empirical software engineering and can serve as strong foundation to generalize our findings, especially considering that an exhaustive study of all possible CPDP techniques and dataset is unrealistic.

\section{Related Work}
\label{sec:related work}

Software defect prediction is one of the most active research areas in software engineering~\cite{Marco2010,LeeNHKI11,Tim2007,NamPK13,Foyzur2011,Thomas2008}. Generally speaking, the purpose of defect prediction is to learn an accurate model (supervised or unsupervised) from a corpus of data (e.g., static code features, churn data, defect information) and apply the model to new and unseen data. To this end, the data can be labeled by using code metrics~\cite{Akiyama71,McCabe76,ChidamberK94,AbreuC94}; process metrics~\cite{NagappanB05,MoserPS08,Hassan09,LeeNHKI11}; or metrics derived from domain knowledge~\cite{MeneelyWSO08,TabaKZHN13}. 
Depending on the scenario, the training data can come from the same project that one aims to predict the defects for, i.e., within project defect prediction or from other projects, i.e., CPDP. For supervised learning, CPDP consists of two parts: domain adaptation and classification where the former is resolved by transfer learning while the latter is tackled by a machine learning classifier.

In the past two decades, CPDP has attracted an increasing attention, as evidenced by many survey work~\cite{HerboldTG18,Herbold17,HosseiniTG19,ZhouYLCLZQX18}. In CPDP, the homogeneous CPDP problem, which we focus on this work, refers to the case where the metrics of the source and target projects are the exactly the same or at least contain the same ones. Among others, instance selection is the earliest and most common way to transfer the learned model for CPDP, in which similar source instances to the target instances are selected to learn a model~\cite{TurhanMBS09,PetersMM13,7336025,RyuJB15,AmasakiKY15}. Alternatively, instance weighting uses different weights for each source instance, depending on the relevance of the instance to the target instances, see \cite{MaLZC12}, \cite{ChenFST15} and \cite{RyuCB16}. Projects and feature selection is another way to transfer the learned a model when there are multiple source projects~\cite{HePMY13,HeSYLW12}, \cite{HosseiniTM16,Herbold13,HeLLCM15}. Finally, instance standardization exist for CPDP, in which the key idea is to transform source and target instance into a similar form (e.g., distribution and representation)~\cite{NamPK13,ZhangMKZ16}. More comprehensive summaries about techniques for CPDP can be found in the survey by Zimmermann et al.~\cite{ZimmermannNGGM09} and He et al.~\cite{HeSYLW12}.

More recently, studies have shown that CPDP can be improved by using different models \cite{Cao2015SoftwareDP}, model combination \cite{XiaLPNW16}\cite{PanichellaOL14}\cite{HerboldTG17} or a model ensemble \cite{UchigakiUTM12}\cite{ZhangLXS15}. Another way to improve prediction performance is via data preprocessing before applying a CPDP techniques~\cite{Watanabe08}\cite{CruzO09}\cite{abs-1901-08429}, or directly using an unsupervised learning have, such as the work by Nam and Kim~\cite{NamK15}.

Despite the tremendous amount of studies on the defect prediction models and approaches to improve the prediction performance, their parameter optimization has not received enough attentions. This is in fact non-trivial, as we have shown that more than 80\% of the defect prediction models have at least one configurable and adaptable parameter. However, most work on defect prediction assumes default settings or relies on ad-hoc methods, which provide little guarantee on the achieved performance. This is an even more serious issue when considering CPDP, in which case the number of parameters and the possible configuration options can be enlarged dramatically. Very few studies have been performed on the automated parameter optimization for defect prediction models. Lessmann et al.~\cite{LessmannBMP08} are among the first to conduct automated parameter optimization for defect prediction models by using grid search. Agrawal et al.~\cite{abs-1902-01838} preform a more recent study and propose an optimization tool called \texttt{DODGE}, which eliminates the need to evaluate some redundant combinations of preprocessors and learners. Nevertheless, unlike our work, these studies neither aim to empirically evaluate the impact of automated parameter optimization nor focus on CPDP.

The most related work is probably the empirical study from Tantithamthavorn et al.~\cite{Tantithamthavorn16}, in which they perform the first thorough study on the impact of automated parameter optimization for defect prediction models \cite{Tantithamthavorn19}. However, our work is fundamentally different from theirs in the following aspects:

\begin{itemize}
    \item \emph{\underline{Considering cross-projects:}} We empirically study the automated parameter optimization for transfer learning based models on cross-project prediction while Tantithamthavorn et al.~\cite{Tantithamthavorn16} focus on the optimization for the defect prediction model within a single project.
    \item \emph{\underline{Studying a larger set of models:}} The number of combinations of transfer learner and classifier considered in our experiments constitutes 62 CPDP techniques. This amount is nearly six times more than the 11 classifiers studied by Tantithamthavorn et al.~\cite{Tantithamthavorn16}.
    \item \emph{\underline{Providing wider insights:}} Our findings, apart from confirming that the automated parameter optimization on CPDP is effective, also provides insights on the algorithm selections for CPDP. In contrast, Tantithamthavorn et al.~\cite{Tantithamthavorn16} mainly provide analysis on the effectiveness and stability of automated parameter optimization.
\end{itemize}

In summary, our work is, to the best of our knowledge, the first comprehensive empirical study about the impact of automated parameter optimization on transfer learning for CPDP, based on which we have revealed important findings and insights that have not been known before~\cite{LiZZL09,LiZLZL09,LiFK11,LiKWCR12,CaoKWL12,LiKCLZS12,LiKWTM13,LiWKC13,LiK14,CaoKWL14,LiZKLW14,LiFKZ14,CaoKWLLK15,WuKZLWL15,LiDZK15,LiKZD15,LiKD15,LiDZ15,LiDZZ17,WuKJLZ17,WuLKZZ17,LiDAY17,LiDY18,ChenLY18,WuLKZ20,ChenLBY18,LiCFY19,WuLKZZ19,LiCSY19,Li19,BillingsleyLMMG19,ZouJYZZL19,LiuLC19,LiXT19,GaoNL19}.

\section{Conclusions}
\label{sec:conclusions}

In this paper, we conduct the first empirical study, which offers an in-depth understanding on the impacts of automated parameter optimization for CPDP based on 62 CPDP techniques across 20 real-world projects. Our results reveal that:

\begin{itemize}
    \item Automated parameter optimization can significantly improve the CPDP techniques. Up to 77\% of the improvement exhibits \emph{huge} effect size under the Cohen's rule.
   \item Optimizing the parameters of transfer learning techniques plays a more important role in performance improvement in CPDP.
   \item The state-of-the-arts combinations of transfer learning and classification are far from mature, as the statistically best technique comes from the 49 \lq newly\rq\ developed combinations in most cases.
\end{itemize}
Our findings provide valuable insights for the practitioners from this particular research field to consider. Drawing on such, in our future work, we will design sophisticated optimizer for CPDP that explicitly searches the parameter space for the transfer learning part. Furthermore, the problem of portfolio optimization for CPDP, which involves both the selection of combination and parameter tuning, is also one of our ongoing research directions.

%on the performance of CPDP techniques by conducting a series of empirical experiments. First, we study whether automated parameter optimization can improve the performance of existing CPDP techniques. And we find that automated parameter optimization is indeed effective to improve the performance of most CPDP techniques. Then we compare the effectiveness of the different types of optimization. From the result, we can find that domain adaptation is vital for CPDP models while few works focus on optimizing domain adaptation. Finally, we find that the settings of existing CPDP techniques are not optimal. The results suggest that researchers should pay more attention to the settings of transfer techniques and not just use default settings. 

%CPDP is basically composed of two parts: domain adaptation and classification, therefore the parameter optimization has to take two parts into account. In this paper, we only study the parameter optimization of a given combination. How to choose the best combination of transfer technique and classifier will be studied in our future work. And in terms of the different types of optimization, we will make a more proper comparison in the future.

\section*{Acknowledgement}
K. Li was supported by UKRI Future Leaders Fellowship (Grant No. MR/S017062/1) and Royal Society (Grant No. IEC/NSFC/170243).

\bibliographystyle{IEEEtran}
\bibliography{IEEEabrv,sample-base}

\end{document}